\def\paperID{14556}
\def\confName{CVPR}
\def\confYear{2025}

\def\authorBlock{
    Zefeng Zhang\textsuperscript{1, 2} \qquad
    Hengzhu Tang\textsuperscript{3} \qquad
    Jiawei Sheng\textsuperscript{1, 2} \qquad
    Zhenyu Zhang\textsuperscript{3}\thanks{Project lead.}\qquad \\
    Yiming Ren\textsuperscript{1, 2} \qquad
    Zhenyang Li\textsuperscript{3} \qquad
    Dawei Yin\textsuperscript{3} \qquad
    Duohe Ma\textsuperscript{1, 2†} \qquad
    Tingwen Liu\textsuperscript{1, 2}\thanks{Corresponding author.} \\

    \textsuperscript{1}Institute of Information Engineering, Chinese Academy of Sciences \\
    \textsuperscript{2}School of Cyber Security, University of Chinese Academy of Sciences,\quad 
    \textsuperscript{3}Baidu Inc. \\
    {\tt\small \{zhangzefeng, shengjiawei, renyiming, maduohe, liutingwen\}@iie.ac.cn} \\
    {\tt\small \{tanghengzhu, zhangzhenyu07, zhenyounglee, yindawei\}@baidu.com}
}

\newif\ifreview 
\newif\ifarxiv 
\newif\ifcamera \newcommand{\cameraready}{\cameratrue}
\newif\ifrebuttal 

\cameraready

\pdfoutput=1
\documentclass[10pt,twocolumn,letterpaper]{article}

\usepackage[accsupp]{axessibility} 

\ifreview \usepackage[review]{cvpr} \fi
\ifarxiv \usepackage[pagenumbers]{cvpr} \fi
\ifrebuttal \usepackage[rebuttal]{cvpr} \fi
\ifcamera \usepackage{cvpr} \fi

\usepackage{graphicx}	
\usepackage{amsmath}	
\usepackage{amssymb}	
\usepackage{booktabs}
\usepackage{times}
\usepackage{microtype}
\usepackage{epsfig}
\usepackage{caption}
\usepackage{float}
\usepackage{placeins}
\usepackage{color, colortbl}
\usepackage{stfloats}
\usepackage{enumitem}
\usepackage{tabularx}
\usepackage{xstring}
\usepackage{multirow}
\usepackage{xspace}
\usepackage{url}
\usepackage{subcaption}
\usepackage{xcolor}
\usepackage[hang,flushmargin]{footmisc}

\ifcamera \usepackage[accsupp]{axessibility} \fi

% ...

\ifarxiv  \fi

\newcommand{\R}[1]{{%
    \textbf{%
        \ifstrequal{#1}{1}{\textcolor{red}{R#1}}{%
        \ifstrequal{#1}{2}{\textcolor{blue}{R#1}}{%
        \ifstrequal{#1}{3}{\textcolor{magenta}{R#1}}{%
        \ifstrequal{#1}{4}{\textcolor{teal}{R#1}}{%
                           \textcolor{cyan}{R#1}%
        }}}}%
    }%
}}

\usepackage{xr-hyper}

\makeatletter
\newcommand*{\addFileDependency}[1]{
  \typeout{(#1)}
  \@addtofilelist{#1}
  \IfFileExists{#1}{}{\typeout{No file #1.}}
}

\makeatother
\newcommand*{\myexternaldocument}[1]{
    \externaldocument{#1}
    \addFileDependency{#1.tex}
    \addFileDependency{#1.aux}
}

\definecolor{cvprblue}{rgb}{0.21,0.49,0.74}
\usepackage[pagebackref,breaklinks,colorlinks,allcolors=cvprblue]{hyperref}
\usepackage[capitalize]{cleveref}
\crefname{section}{Sec.}{Secs.}
\crefname{table}{Table}{Tables}
\crefname{figure}{Fig.}{Figs.}

\ifarxiv \crefname{appendix}{App.}{Apps.}
\else \crefname{appendix}{Suppl.}{Suppls.} \fi

\unless\ifarxiv \myexternaldocument{_supplementary} \fi

\begin{document}
\title{Debiasing Multimodal Large Language Models via \\ Noise-Aware Preference Optimization}
\author{\authorBlock}
\maketitle

\begin{abstract}

Multimodal Large Language Models (MLLMs) excel in various tasks, yet often struggle with modality bias, where the model tends to rely heavily on a single modality and overlook critical information in other modalities, which leads to incorrect focus and generating irrelevant responses. 
In this paper, we propose using the paradigm of preference optimization to solve the modality bias problem, including RLAIF-V-Bias, a debiased preference optimization dataset, and a Noise-Aware Preference Optimization (NaPO) algorithm. 
Specifically, we first construct the dataset by introducing perturbations to reduce the informational content of certain modalities, compelling the model to rely on a specific modality when generating negative responses. 
To address the inevitable noise in automatically constructed data, we combine the noise-robust Mean Absolute Error (MAE) with the Binary Cross-Entropy (BCE) in Direct Preference Optimization (DPO) by a negative Box-Cox transformation, and dynamically adjust the algorithm’s noise robustness based on the evaluated noise levels in the data.
Extensive experiments validate our approach, demonstrating not only its effectiveness in mitigating modality bias but also its significant role in minimizing hallucinations.
The code and data is available at 
\hyperlink{https://github.com/zhangzef/NaPO}{https://github.com/zhangzef/NaPO}.


\end{abstract}

\section{Introduction}
\label{sec:intro}

\begin{figure}
\centering
\includegraphics[width=0.48 \textwidth,height=0.35\textwidth]{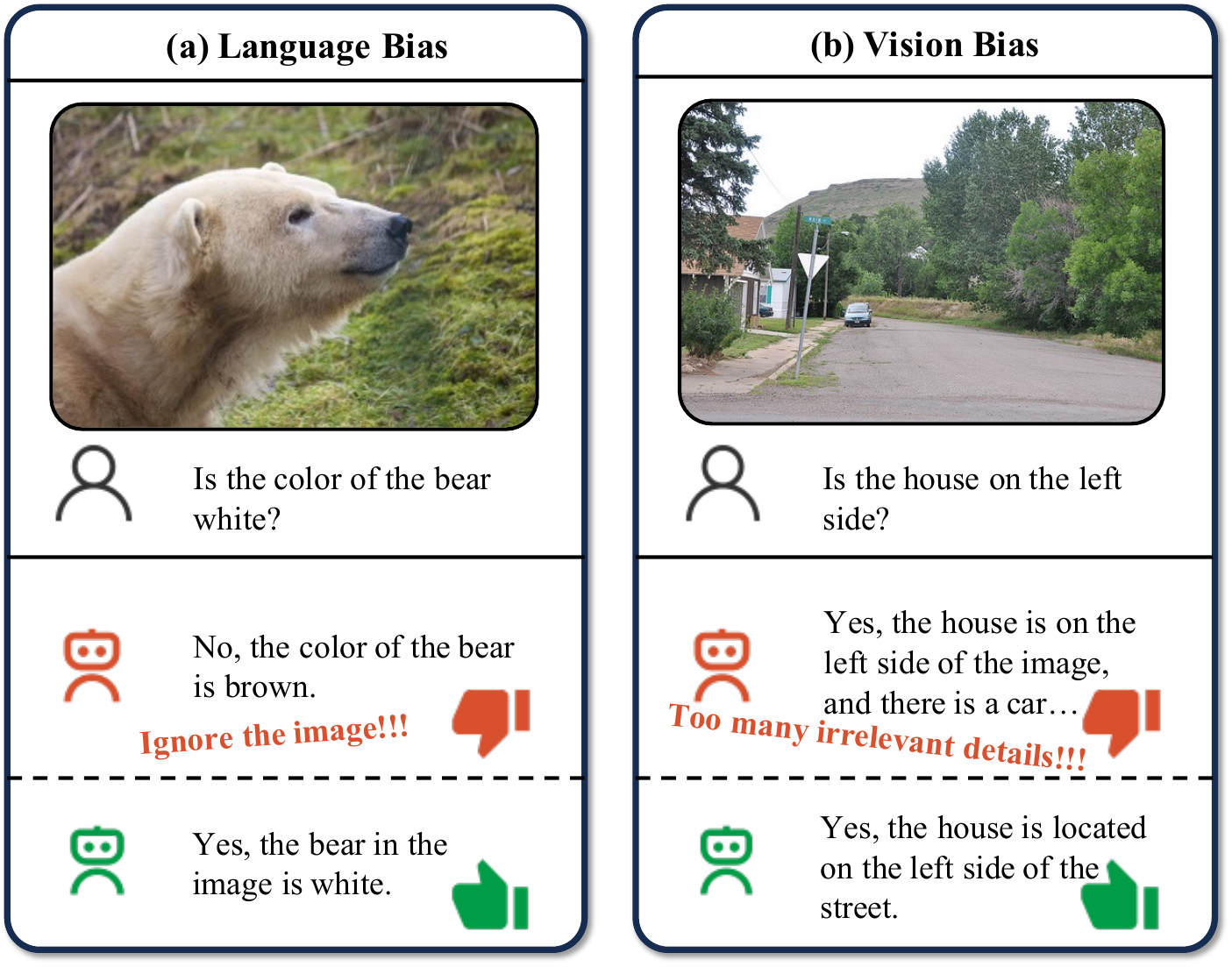}
\caption{Examples of different types of modality-biased responses and their preferred counterparts. \textbf{Left}: The model relies excessively on prior knowledge, assuming a bear is brown while \textit{overlooking the image}, which shows a polar bear. \textbf{Right}: Although the model answers the question correctly, it provides \textit{unnecessary image details} that are irrelevant to the question.} 
\label{fig:modalityBias}
\end{figure}

Multimodal Large Language Models (MLLMs) have achieved powerful multimodal understanding capabilities~\cite{zhang2024optimal, zhang2024revealing, tang2025multi} through pretraining on large-scale image-text data, significantly advancing AI research~\cite{achiam2023gpt, li2023blip, liu2024visual, zhu2023minigpt, ye2023mplug, bai2023qwen, liu2024improved}. 
These MLLMs fuse large-scale pre-trained vision models into the representation space of the Large Language Models (LLMs), allowing the LLMs access to the visual representations.
However, MLLMs continue to struggle with the \textit{modality bias}~\cite{lee2024vlind, chen2024quantifying, zhang2024debiasing}, where the model tends to rely heavily on one of the involved modalities and overlook critical information from other modalities, leading to incorrect focuses and irrelevant responses.
Specifically, in models with both text and image inputs, modality biases are mainly manifested as the \textit{language bias} and \textit{vision bias}. 
Figure~\ref{fig:modalityBias}~(a) shows an example of the \textit{language bias}: for the question “Is the color of the bear white?”, the model overly relies on the language priors~\cite{zhu2020overcoming} that most bears are brown, and overlooks the critical information of “polar bear” with white color in the input image.
This bias inevitably leads to incorrect responses and would be even worse when larger LLMs dominate MLLMs.
Besides, Figure~\ref{fig:modalityBias}~(b) shows an example of the \textit{vision bias}: when asking “Is the house on the left side?”, the model focuses too much on image details, resulting in a lack of accurate understanding of the textual question.
Such modality bias would add trivial irrelevant information in model responses, exacerbating the challenge of instruct following~\cite{instructfollowing}.

An ideal MLLM ought to be modality-unbiased, effectively integrating information from all modalities to provide accurate and complete answers~\cite{zhang2024debiasing}. 
Existing studies have attempted balanced dataset distribution for training~\cite{gokhale2020mutant,DBLP:conf/wacv/KollingMGPPB22}, or devise strategies to identifying and mitigating bias during training or inference~\cite{DBLP:conf/emnlp/WangLW21,DBLP:conf/aaai/HuangQQSZ22,DBLP:conf/emnlp/PatilMB23}. 
However, these studies mostly require additional large-scale supervised fine-tuning, which risks losing valuable existing knowledge in MLLMs~\cite{DBLP:conf/emnlp/DongDSXSL22,DBLP:conf/mm/WangLQLTCS24}.
In contrast, we notice that debiasing MLLMs can be seen as a \textit{preference optimization} problem~\cite{ouyang2022training} in LLMs, which can enhance alignment with human preferences by increasing the probability gap between preferred (unbiased) and non-preferred (biased) generated responses. 
In other words, by increasing the generation probability of unbiased responses over biased ones, the model is expected to incorporate critical information from all modalities, thus alleviating the bias on a single modality.
In this way, this debiasing design adjusts the model preference on response generation, yet remains most existing knowledge in MLLMs with original capabilities.
Additionally, it is still challenging to derive high-quality datasets for preference optimization. 
To our knowledge, there are few preference optimization datasets specifically for MLLM debiasing, and automatically constructed datasets often contain significant noise since biased responses, though present, are not always of low quality. 
This makes standard preference optimization algorithms struggle when facing potentially noisy (incorrect) unbiased and biased preference optimization data.


To this end, starting from the queries in RLAIF-V~\cite{yu2024rlaif}, we design a data construction method to generate biased data by perturbing other modalities to prompt the model to rely excessively on a single modality and produce biased responses. 
Specifically, we generate language-biased and vision-biased responses by selectively masking visual and textual information in the input, attach such biased responses to RLAIF-V  as negative samples, and finally achieve a new preference optimization dataset with modality bias, termed RLAIF-V-Bias.
Besides, we propose introducing Noise-Aware Preference Optimization (NaPO) to dynamically identify noisy data and reduce optimization weights for these samples. 
Specifically, NaPO builds on Generalized Cross Entropy~\cite{zhang2018generalized} by applying a negative Box-Cox transformation~\cite{box1964analysis}, combining noise-robust Mean Absolute Error (MAE)\cite{ghosh2015making} with Binary Cross-Entropy (BCE) from Direct Preference Optimization (DPO)\footnote{See Section\ref{sec:noise_robustness_analysis} for a detailed comparison of the two loss functions.}. 
NaPO’s noise robustness coefficient is dynamically adjusted by assessing the noise level of training samples.

To evaluate effectiveness, we take LLaVA-v1.5-7b~\cite{liu2024improved} as the base model to generate negative samples and perform preference optimizations, following the setup of RLAIF-V~\cite{yu2024rlaif}.
Next, we evaluate the proposed RLAIF-V-Bias dataset and the NaPO algorithm on VLind-Bench~\cite{lee2024vlind} (a benchmark for language priors and commonsense biases in MLLMs) as well as on common hallucination benchmarks: Object HalBench~\cite{rohrbach2018object}, MMHalBench~\cite{sun2023aligning}, and AMBER~\cite{wang2023llm}. 
Compared to the original training set and DPO, our approach showed approximately 19.5\% and 18.6\% improvements in reducing bias and language priors, with a notable reduction in hallucinations relative to the baseline. 
We also tested the effectiveness of our method and dataset on models with varying parameter sizes.

Our contributions are threefold: 
First, we develop a data construction method based on the causes of modality bias and practice it to create a debiasing-oriented preference optimization dataset RLAIF-V-Bias.
Second, we propose NaPO, which applies the negative Box-Cox transformation to DPO, enabling it to adjust the loss function’s noise robustness, and we further design a dynamic noise assessment method that allows NaPO to adapt its noise robustness dynamically during training based on data analysis.
Lastly, we demonstrate through experiments on language-prior and hallucination benchmarks that our method effectively mitigates the modality bias problem in MLLMs.

\section{Preliminaries}
\label{sec:preliminaries}

\subsection{Preference Optimization}
Preference optimization aims to align LLMs with human preferences, thereby enhancing their responsiveness to human needs.
The RL-based preference optimization method first trains a Supervised Fine-Tuning (SFT) model using human-labeled preference data to obtain a reward model. 
Then, it simulates the environment through the reward model and uses algorithms such as Proximal Policy Optimization (PPO) to maximize the LMs' reward, achieving alignment with human preferences~\cite{christiano2017deep}.
Due to the complexity and high resource consumption of RL-based methods, many studies focus on designing simplified loss functions that enable LLMs to align with human preferences directly using this loss function and human-labeled preference data. 
DPO~\cite{rafailov2024direct} is one of the promising approaches. 
Given an input $x$ and a response $y$, DPO defines its reward $r(x,y)$ as:
\begin{equation}
    r(x,y) = \beta \log \frac{\pi_{\theta}(y|x)}{\pi_{ref}(y|x)} + \beta \log Z(x),
\end{equation}
where $Z(x)$ is a partition function, $\beta$ is a hyperparameter that controls the deviation from the reference model, $\pi_{\theta}$ and $\pi_{ref}$ denote the policy model and the reference model, respectively.
Given a preference optimization sample $(x, y_w, y_l)$, where $y_w$ denotes the preferred response and $y_l$ the rejected response, DPO aligns the LLM with human values by maximizing the reward margin between $y_w$ and $y_l$ based on the Bradley-Terry model~\cite{bradley1952rank}:
\begin{equation} \label{eq:reward_margin}
\begin{cases}
    \psi_{\Sigma}(x, y_w, y_l) = \beta \log \frac{\pi_{\theta}(y_w|x)}{\pi_{ref}(y_w|x)} - \beta \log \frac{\pi_{\theta}(y_l|x)}{\pi_{ref}(y_l|x)}, \\
    \psi_{\mu}(x, y_w, y_l) = \frac{\beta}{|y_w|} \log \frac{\pi_{\theta}(y_w|x)}{\pi_{ref}(y_w|x)} - \frac{\beta}{|y_l|} \log \frac{\pi_{\theta}(y_l|x)}{\pi_{ref}(y_l|x)}, \\
\end{cases}
\end{equation}
where $\psi_{\Sigma}$ denotes calculating the reward margin using the sum of log probabilities (logP), while $\psi_{\mu}$ denotes calculating the reward margin using the average logP. $|y|$ denotes the length of the token sequence.
With the default setting of $\psi_{\Sigma}$, DPO uses a BCE loss to enhance the reward difference between the $y_w$ and $y_l$ for LLMs:
\begin{equation} \label{eq:DPO}
    \mathcal{L}_{\text{DPO}} = - \log \sigma \bigg(\beta \log \frac{\pi_{\theta}(y_w|x)}{\pi_{ref}(y_w|x)} - \beta \log \frac{\pi_{\theta}(y_l|x)}{\pi_{ref}(y_l|x)} \bigg).
\end{equation}

\subsection{Noise Robustness Analysis}
\label{sec:noise_robustness_analysis}
A loss function is considered noise-robust if it minimizes risk similarly with both noisy and noise-free labels~\cite{ghosh2015making}. This means the loss function can suppress or reduce the impact of noisy data on the optimization process.
As a commonly used classification loss, BCE converges quickly but is prone to overfitting on noisy data, whereas MAE is noise-robust but converges slowly, which can result in undertrained models.
Here we compare these two loss functions from the perspectives of symmetric loss and gradients.
Now, let’s consider a simple binary classification scenario. For a training sample $\{x, y\}$, $y \in \{0, 1\}$, MAE and BCE can be formalized as:
\begin{equation}
\begin{cases} 
\mathcal{L}_{\text{MAE}} = |y - f(x)|, \\
\mathcal{L}_{\text{BCE}} = y\log(f(x)) + (1-y)\log(1-f(x)).
\end{cases}
\end{equation}
For the first perspective, \citet{ghosh2015making} have proven that symmetric loss functions exhibit superior noise robustness.
A loss function $\mathcal{L}$ is a symmetric loss if and only if it satisfies the following equation for any $x$ and classifier $f$~\cite{ghosh2015making}:
\begin{equation} \label{eq:sym}
    \mathcal{L}(f(x), y) + \mathcal{L}(f(x), 1-y) = C,
\end{equation}
where $C$ is a constant. When the loss function $\mathcal{L}$ is MAE, $C=1$. However, when the loss function $\mathcal{L}$ is BCE, the result of the Equation~(\ref{eq:sym}) is  $- \left( \log(f(x)) + \log(1 - f(x)) \right)$ , which is not a constant. Therefore, MAE has better noise robustness compared to BCE.
From the perspective of gradients, the gradients of the two are as follows:
\begin{equation}
    \begin{aligned}
        \frac{\partial \mathcal{L}(f_{\theta}(x), y)}{\partial \theta} = 
\begin{cases}
-\frac{1}{f_{\theta}(x)} \nabla_{\theta} f_{\theta}(x) & \text{for BCE},\\
-\nabla_{\theta} f_{\theta}(x) & \text{for MAE}.
\end{cases}
    \end{aligned}
\end{equation}
So the smaller $f_{\theta}(x)$ or larger $\frac{1}{f_{\theta}(x)}$, are implicitly weighed more than samples with predictions that agree more with provided labels in the gradient update. 
This implies that in BCE training, greater emphasis is placed on harder samples, helping the model to converge quickly but potentially causing it to overfit on noisy data.
In summary, while BCE \textit{converges quickly}, it tends to overfit on noisy data, whereas MAE is \textit{noise-robust} but converges more slowly, which may lead to suboptimal training performance~\cite{zhang2018generalized}.

\section{Method}
\label{sec:method}

\begin{figure*}
\centering
\includegraphics[width=1\textwidth]{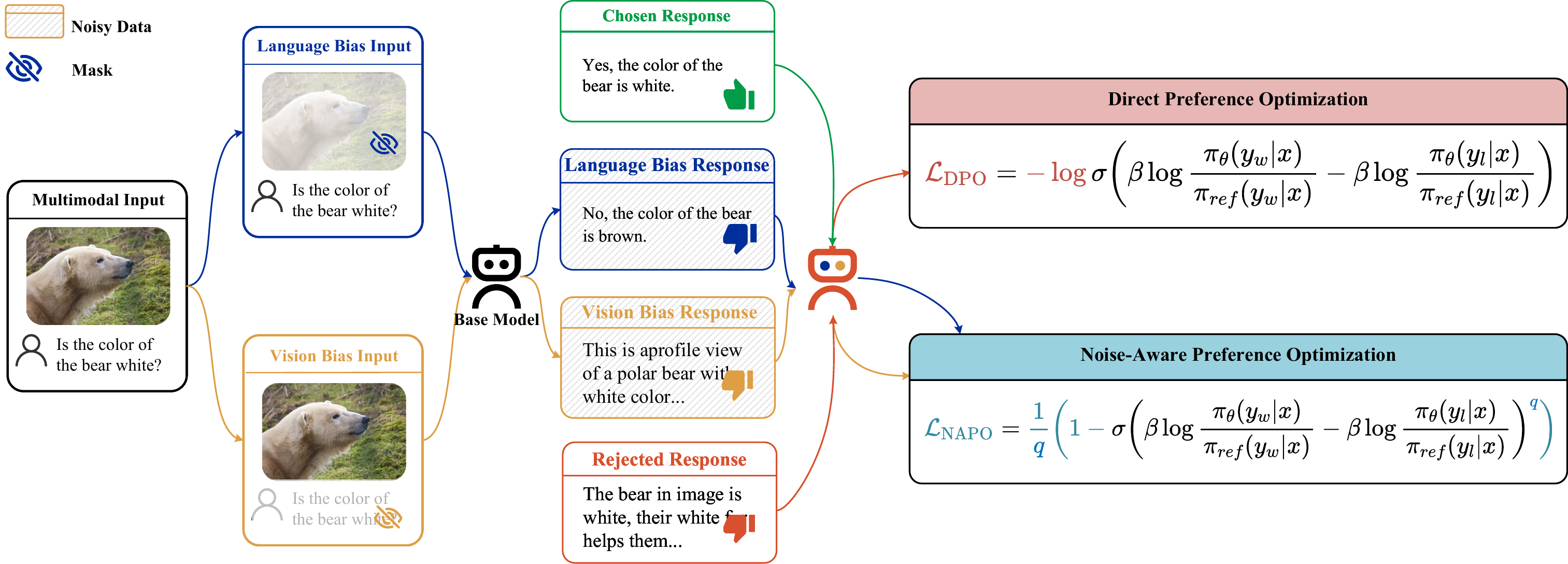}
\caption{\textbf{Method details.} First, biased responses are constructed by using masking to guide the model toward over-relying on prompts and generating responses based on the base model. Next, NaPO is applied for noise-robust preference optimization to counteract noise in automatically constructed data, dynamically assessing data noise levels to calculate NaPO’s noise robustness coefficient $q$ (see Equation~(\ref{eq:dynq})). Here we assumed that the original data is of high quality, so DPO is used to train on it directly. Additional experiments were conducted with NaPO on the original data, and the results can be found in Appendix~\ref{sec:additional_experiments}.} 
\label{fig:model}
\end{figure*}

This section details modality bias mitigation in MLLMs within the Bradley-Terry preference optimization framework, achieved by reducing biased response probabilities and increasing ground truth likelihoods. 
The approach consists of two main components: biased response generation and noise-aware preference optimization. 
In Section~\ref{sec: bias_gen}, we describe the biased response generation process, while Section~\ref{sec:napo} covers noise-aware preference optimization and includes an analysis of data for noise distribution to guide bias reduction.

\subsection{Modality Biase Response Generation}
\label{sec: bias_gen}

A biased response is one where the model disproportionately relies on a single modality. 
To encourage this, we introduce controlled disturbances to other modalities, reducing their informational weight and increasing reliance on the target modality. 
We generate biased responses using the RLAIF-V dataset~\cite{yu2024rlaif}, a multimodal preference optimization dataset built through iterative feedback from open-source models.

\paragraph{Language-biased response generation.}
Language bias refers to the model’s excessive reliance on prior knowledge or textual information when generating answers. For a multimodal input $x = (v, t)$, where $v$ is the visual information and $t$ is the instruction and textual context, we reduce the model’s reliance on the visual modality to encourage language-biased responses. Specifically, by masking out all visual information, we guide the model to produce responses that depend primarily on textual information and prior knowledge:
\begin{equation}
    y_{lb} = MLLM([MASK]; t),
\end{equation}
where $y_{lb}$ is the language-biased response, $[MASK]$ denotes the mask tokens used to mask all visual information.

\paragraph{Vision-biased response generation.}
Visual bias occurs when the model focuses too heavily on visual information, producing irrelevant image details in its responses. 
For a multimodal input $(v, t)$, we reduce the model’s reliance on textual input to encourage vision-biased responses. 
Similar to the language-biased response generation, by masking out all textual information, we guide the model to generate responses that depend mainly on visual information:
\begin{equation}
    y_{vb} = MLLM(v; [MASK]),
\end{equation}
where $y_{vb}$ refers to vision-biased response, and $[MASK]$ denotes the mask tokens used to mask all textual information.

\subsection{Noise-Aware Preference Optimization}
\label{sec:napo}

Automatically constructed data inevitably contain noise, allowing some unbiased responses to appear among the generated biased ones, which adversely affects model training. 
Standard preference optimization methods, however, lack robustness to noise and incorporating noise robustness often compromises model convergence~\cite{zhang2018generalized}. 
In this section, we propose a NaPO algorithm.
We first start by combining the BCE from DPO with the noise-robust MAE using a negative Box-Cox transformation. Then, based on data analysis, we introduce an adaptive noise-aware method to dynamically adjust the noise robustness coefficient in NaPO.

\begin{figure}
\centering
\includegraphics[width=0.38 \textwidth,height=0.28\textwidth]{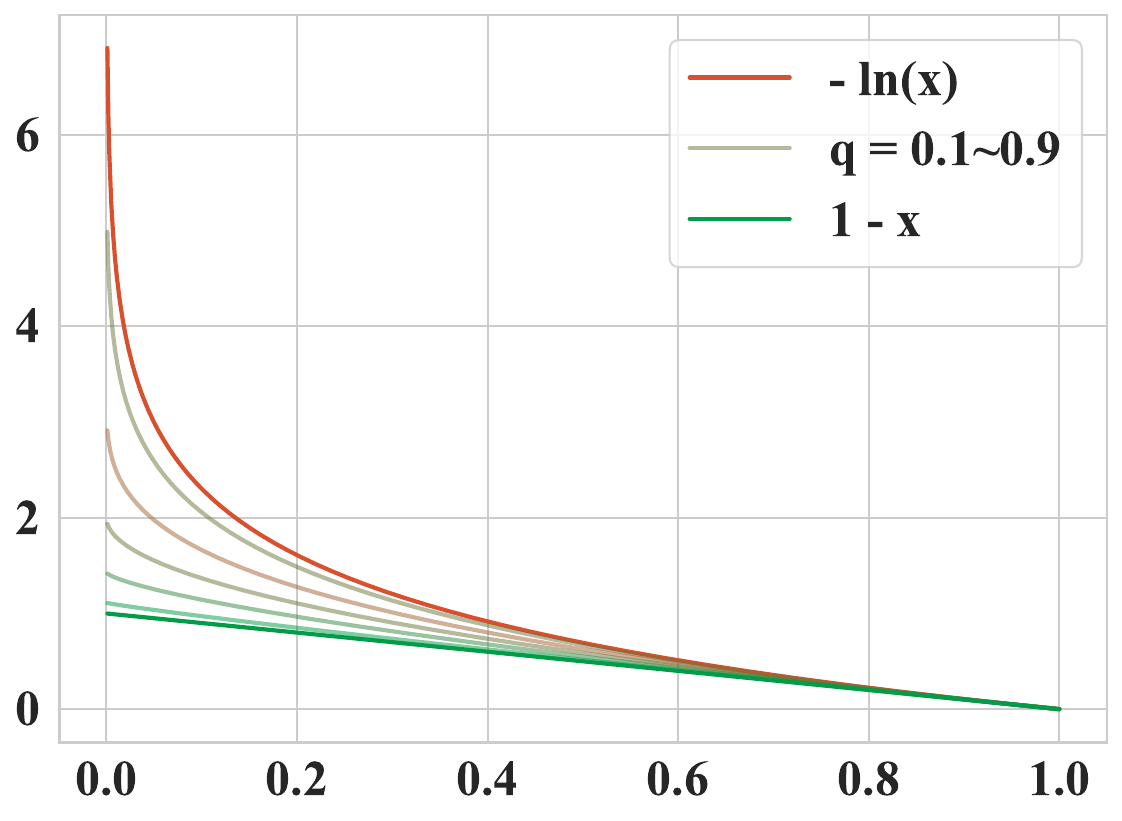}
\caption{\textbf{Comparison of different loss functions.} We plotted the function $(1-x^q)q^{-1}$ for values of $q$ in the range $(0.1, 0.3, 0.5, 0.7, 0.9)$, and compared it with both MAE $(1 - x)$ and BCE $-ln(x)$. By adjusting the value of q, we can balance the noise robustness and the rapid convergence ability of NaPO.} 
\label{fig:functions}
\end{figure}

For a given preference optimization training sample $(x, y_w, y_l)$, DPO uses BCE loss to fit the reward margin activated by the sigmoid function to $1$, enhancing the reward difference between the preferred sample $y_w$ and the non-preferred sample $y_l$. 
As shown in Section~\ref{sec:noise_robustness_analysis}, BCE converges quickly but can be prone to overfitting, whereas MAE is highly noise-robust but converges more slowly. 
To combine these advantages, we use the negative Box-Cox transformation as the loss to achieve both fast convergence (from BCE) and noise robustness (from MAE):
\begin{equation}
\label{eq:NaPO}
\begin{aligned}
\mathcal{L}_{\text{NaPO}} (x, y_w, y_l) = \frac{1}{q} \bigg( 1 - \sigma \bigg( 
& \beta \log \frac{\pi_\theta(y_w | x)}{\pi_{\text{ref}}(y_w | x)} \\
& - \beta\log \frac{\pi_\theta(y_l | x)}{\pi_{\text{ref}}(y_l | x)} 
\bigg)^q \bigg),
\end{aligned}
\end{equation}
where  $q \in (0, 1]$ is the noise robustness coefficient. 
By L’Hôpital’s rule, as $q$ approaches $0$, the function’s limit converges to the BCE:
\begin{equation}
    \lim_{q \to 0} \frac{1}{q}(1 - x^q) = -log(x).
\end{equation}
Therefore, when $q \in (0,1]$, the range of values for $\mathcal{L}_{\text{NaPO}}$ is:
\begin{equation}
    \mathcal{L}_{\text{MAE}} \leq \mathcal{L}_{\text{NaPO}} < \mathcal{L}_{\text{BCE}}.
\end{equation}
We can observe the differences between the various loss functions in Figure~\ref{fig:functions}.
As the value of $q$ increases, the NaPO becomes more similar to MAE, resulting in stronger noise robustness but slower convergence. Conversely, as $q$ decreases, NaPO approaches BCE, which reduces noise robustness but accelerates convergence.
This also indicates that the upper bound of NaPO is DPO.

\paragraph{Adaptive noise robustness coefficient ($q$).}
The long-tail issue in LLMs' knowledge~\cite{kandpal2023large} causes varying degrees of biases in responses to common and rare questions, making a fixed noise robustness coefficient insufficient and less flexible. 
For instance, the model may answer correctly about a brown bear’s color without visual input (coincidentally unbiased response), and produce an incorrect response when asked about a polar bear (biased response). 
In this section, we would like to design a dynamic noise robustness coefficient for biased samples based on data observations.

\begin{figure}
\centering
\includegraphics[width=0.48 \textwidth,height=0.38\textwidth]{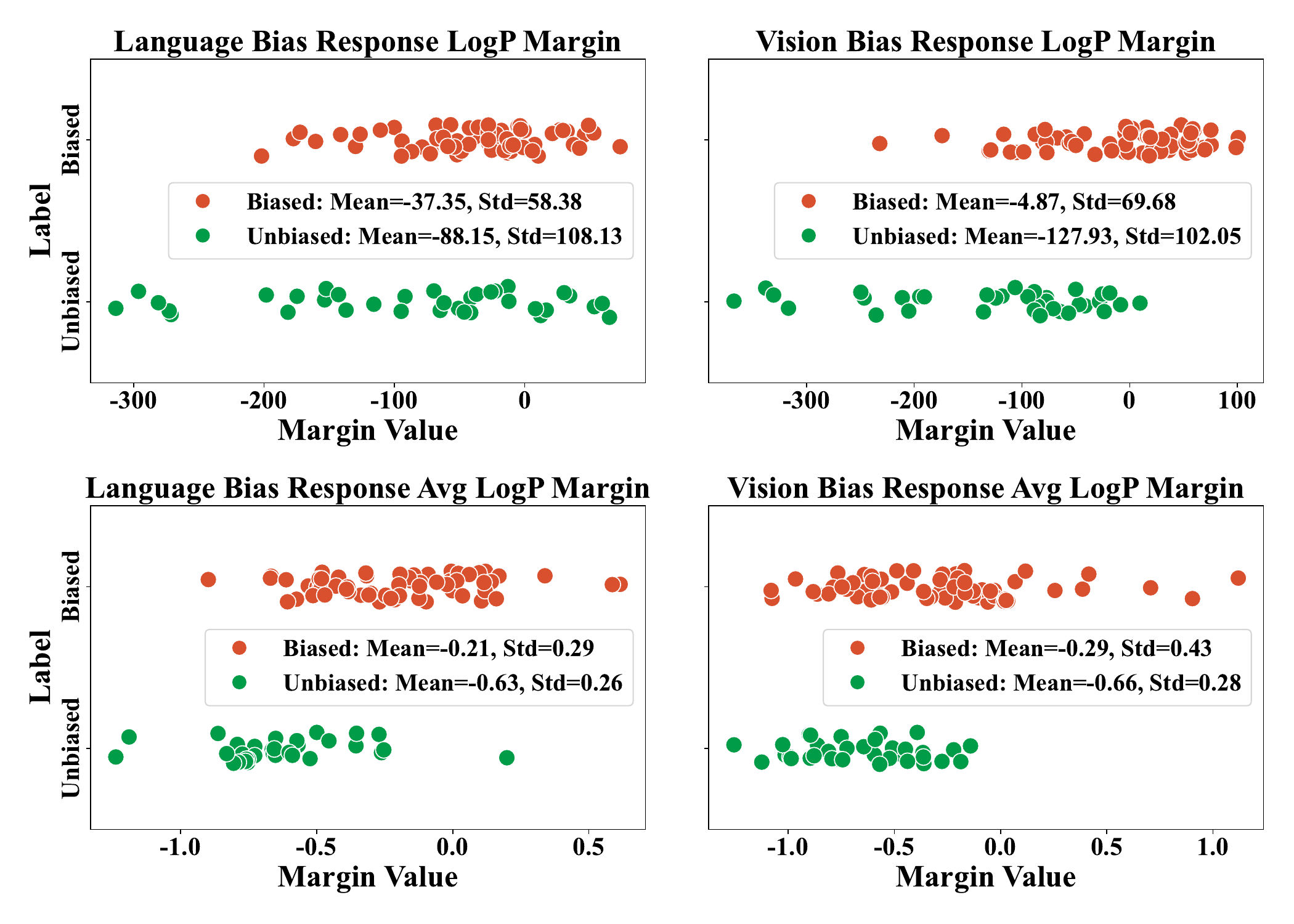}
\caption{\textbf{Analysis of noise and margin distribution in automatically constructed data.} 	The first row shows LogP margins between each biased response type and the ground truth, while the second row shows avg LogP margins. In language-biased responses, biased (noise-free) data have a higher avg LogP margin than unbiased (noise) data. Similarly, in vision-biased responses, biased (noise-free) data show a higher LogP margin than unbiased (noise) data.} 
\label{fig:dist}
\end{figure}

For a given input $x$, the sum of log probabilities of the answer $y$ can reflect the LLM’s confidence, where a smaller log probability margin indicates greater similarity to the ground truth. 
We randomly sample 100 data points and manually annotate each automatically generated response as either biased (noise-free) or unbiased (noise) for ease of observation.
Figure~\ref{fig:dist} shows that, for language-biased responses, the avg LogP margin of noise data is generally smaller than that of noise-free data. 
Similarly, in vision-biased responses, the LogP margin for noise data is smaller than that for noise-free data\footnote{A detailed analysis of this phenomenon is provided in the Appendix~\ref{sec:data_observation_and_analysis}, and experimental evidence supporting this conclusion can be found in Table~\ref{tab:margin_analysis}.}. 
Since a larger coefficient $q$ indicates stronger noise robustness, we observe an inverse trend, i.e., \textit{$q$ and the margin value ought to be generally inversely proportional.}
Based on the above observations, we dynamically derive the coefficient $q$ as follows:
\begin{equation}
\label{eq:dynq}
    q = 1-\sigma(\alpha \psi(x, y_w, y_l)),
\end{equation}
where $\psi(x, y_w, y_l)$ is the reward margin formula $\psi$ in Equation~(\ref{eq:reward_margin}), and $\alpha$ is a scaling factor that normalizes the reward margin within the high-gradient range of the sigmoid function for effective reward capture. 
We use the $\psi_{\mu}$ to calculate $q$ for language-biased responses and the $\psi_{\Sigma}$ for vision-biased responses based on the above observation. 
To ensure training stability, we use the batch-average reward margin to calculate $\psi$ and finally derive $q$.

\paragraph{Final optimization objective.}
From Figure~\ref{fig:dist}, we observe that the margin value is positively correlated with data quality: noisy samples typically have smaller margins, while noise-free samples generally have larger margins. Therefore, we introduce a margin-based dynamic weight in the final optimization objective to balance the loss functions:
\begin{equation}\label{eq:dyn_w}
    \gamma_i = \frac{\psi_{\Sigma}(x, y_w, y_i)}{\sum_i \psi_{\Sigma}(x, y_w, y_i)},\quad i \in [y_l;\ y_{lb};\ y_{vb}],
\end{equation}
where $\psi_{\Sigma}(x, y_w, y_i)$ is the calculation of the reward margin using the sum of LogP in Equation~(\ref{eq:reward_margin}).
Based on, we obtain the final optimization objective as follows:
\begin{equation}\label{eq:final objective}
\begin{aligned}
\mathcal{L}_{\gamma} = & \; \gamma_{y_l} \, \mathcal{L}_{\text{DPO}}(x, y_w, y_l) \\
& + \gamma_{y_{lb}} \, \mathcal{L}_{\text{NaPO}}(x, y_w, y_{lb}) \\
& + \gamma_{y_{vb}} \, \mathcal{L}_{\text{NaPO}}(x, y_w, y_{vb}),
\end{aligned}
\end{equation}
where $\mathcal{L}_{\gamma}$ means the weighted sum of the $\mathcal{L}_{\text{NaPO}}$ and $\mathcal{L}_{\text{DPO}}$ with $\gamma$.
In this way, we adaptively adjust the weights of the loss function to balance the optimization of the general preference, language bias, and vision bias.

\section{Experiments}
\label{sec:experiments}

\subsection{Experimental Setup}
In this section, we briefly introduce the implementation details, baselines, and evaluation settings.

\paragraph{Implementation details.}
Following RLAIF-V~\cite{yu2024rlaif}, we use LLaVA-v1.5-7B as the backbone model to construct our training dataset RLAIF-V-Bias, which is an extension of RLAIF-V, containing both the original data from RLAIF-V and additional bias data.
Consistent with RLAIF-V settings, we set $\beta$ to 0.1, used a learning rate of 5e-7, trained for 4 epochs, and set the batch size to 4. On an 8xA100 80GB machine, data construction took 24 hours, and training took 7 hours.
We also tested our method on LLaVA-v1.5-13B, using the same parameters as for LLaVA-v1.5-7B but training for only one epoch.
For NaPO, we calculate the noise robustness coefficient $q$ for language-biased data using the reward margin based on avg logp with $\alpha$ set to 0.5. 
For vision-biased data, we calculate $q$ using the reward margin based on logp, with $\alpha$ set to 0.01. 
For dynamic loss weights, we derive the values of $a$, $b$, and $c$ based on the logp reward margin during training.
In practice, both $q$ and the dynamic loss weight $a, b, c$ are truncated to the range $[0.01, 1]$.

\paragraph{Baseline approaches.}
We primarily compare our method with standard DPO~\cite{rafailov2024direct}. 
Although MDPO~\cite{wang2024mdpo} is designed for hallucination issues, its approach to addressing language bias is similar to ours; thus, we reproduced MDPO using default settings for comparison. 
We also include results from other multimodal LLMs—GPT-4V~\cite{achiam2023gpt}, LLAVAv1.5-13B~\cite{liu2024improved}, POVID~\cite{zhou2024aligning}, OPERA~\cite{huang2024opera}, VCD~\cite{leng2024mitigating}, EOS~\cite{wu2024self}, HA-DPO~\cite{zhao2023beyond}, HALVA~\cite{sarkar2024mitigating}, RLHF-V~\cite{yu2024rlhf}, HSA-DPO~\cite{xiao2024detecting}, Silkie~\cite{li2023silkie}, and RLAIF-V~\cite{yu2024rlaif}—for reference. However, these results are not directly comparable due to differences in base models, preference data, and alignment methods. Notably, RLAIF-V’s training involves iterative feedback from multiple open-source models, making it challenging to reproduce exactly; thus, we present DPO-based results under default settings for reference.

\paragraph{Evaluation benchmarks.}
We tested our method on a benchmark for evaluating language and commonsense bias in MLLMs, as well as on two hallucination-specific evaluation sets for MLLMs.
VLind-Bench~\cite{lee2024vlind} measures language priors and commonsense bias in MLLMs. It has two of the main metrics, Language Prior (LP) and Commonsense Bias (CB), which are used to evaluate the linguistic and visual bias of the model, respectively.
Object HalBench~\cite{rohrbach2018object} is a standard benchmark for object hallucination. Following \citet{yu2024rlaif}, we augment it with eight diverse prompts for 300 instances and report CHAIR scores~\cite{rohrbach2018object} for hallucination rate at the response (CHAIRs) and object (CHAIRi) levels.
AMBER~\cite{wang2023llm} is a multimodal LLM hallucination benchmark with detailed object annotations, focusing on generative tasks with 1K images. Using the official evaluation tool, we report CHAIR score variants, object coverage, hallucinated response rate, and hallucination overlap with human cognition.
We additionally tested on the GPT-4-based hallucination evaluation dataset MMHalBench~\cite{sun2023aligning}, a practical question-answering benchmark with eight question categories and 12 object topics, using GPT-4~\cite{achiam2023gpt} to assess response quality (scored from zero to six) and hallucination rate.

\subsection{Main Results}

\begin{table*}[t]
    \centering\footnotesize
    \setlength{\tabcolsep}{5pt}
    \begin{tabular*}{\textwidth}{@{\extracolsep{\fill}}@{}l|cccccccccc@{}}
    \toprule
    \multirow{2}{*}{Model} & \multicolumn{2}{c}{\textbf{VLindBench}} &
    \multicolumn{2}{c}{\textbf{Object HalBench}} & \multicolumn{4}{c}{\textbf{AMBER}} &
    \multicolumn{2}{c}{\textbf{MMHalBench}} \\
    \cmidrule{2-3}
    \cmidrule{4-5}
    \cmidrule{6-9}
    \cmidrule{10-11}
    ~ & CB $\uparrow$ & LP $\uparrow$ & CHAIRs $\downarrow$ & CHAIRi $\downarrow$ & CHAIRs $\downarrow$ & Cover. $\uparrow$ & HalRate $\downarrow$ & Cog. $\downarrow$ & Score $\uparrow$ & HalRate $\downarrow$ \\
    \midrule
    \midrule
    GPT-4V & 91.1& 75.6 & 13.6 & 7.3 & 4.6 & 67.1 & 30.7 & 2.6 & 3.49 & 0.28 \\
    \midrule
    LLaVA-v1.5-7B & 0.0 & 0.0 & 53.6 & 25.2 & 7.8 & 51.0 & 36.4 & 4.2  & 2.11 & 0.54 \\
    + HACL & - & - & - & - & - & - & - & - & 2.13 & 0.50 \\
    + OPERA & - & - & 45.1 & 22.3 & - & - & - & - & 2.15 & 0.54 \\
    + POVID & - & - & 48.1 & 24.4 & - & - & - & - & 2.08 & 0.56 \\
    + VCD & - & - & 48.8 & 24.3 & - & - & - & - & 2.12 & 0.54 \\
    + EOS & - & - & 40.3 & 17.8 & 5.1 & 49.1 & 22.7 & 2.0 & 2.03 & 0.59 \\
    + HA-DPO & - & - & 39.9 & 19.9 & 6.7 & 49.8 & 30.9 & 3.3 & 1.97 & 0.60 \\
    + HALVA & - & - & - & - & 6.6 & 53.0 & 32.2 & 3.4 & 2.25 & 0.54 \\
    + MDPO-10K & - & - & 35.7 & 9.8 & 4.4 & 52.4 & 24.5 & 2.4 & 2.39 & 0.54 \\
    + RLAIF-V-Iterative & 54.3 & 35.3 & 8.5 & 4.3 & - & - & - & - & 3.06 & 0.29 \\
    \midrule
    LLaVA-v1.5-13B & 31.5 & 20.9 & 53.3 & 14.5 & 8.5 & 50.9 & 37.6 & 4.2 & 3.03 & 0.47 \\
    + RLHF-V & - & - & 12.2 & 7.5 & 6.3 & 46.1 & 25.1 & 2.1 & 2.81 & 0.49 \\
    + HSA-DPO & - & - & 5.2 & 3.2 & 2.1 & 47.3 & 13.4 & 1.2 & 2.61 & 0.48 \\
    + HALVA & - & - & - & - & 6.4 & 52.6 & 30.4 & 3.2 & 2.58 & 0.45 \\
    \midrule
    \midrule
    LLaVA-v1.5-7B & 0.0 & 0.0 & 53.6 & 25.2 & 7.8 & 51.0 & 36.4 & 4.2 & 2.11 & 0.54 \\
    + $\mathcal{L}_{\text{DPO}}$ with RLAIF-V & 39.4 & 25.4 & 32.0 & 8.5 & 4.9 & 52.0 & 23.4 & 1.6 & 3.23 & \underline{0.38} \\
    + $\mathcal{L}_{\text{MDPO}}$ with RLAIF-V & 0.3 & 0.4 & 35.3 & 10.5 & \underline{4.2} & \underline{53.1} & \underline{22.4} & 2.2  & \underline{3.28} & 0.42 \\
    + $\mathcal{L}_{\gamma}$ with RLAIF-V-Bias & \textbf{58.9} & \textbf{44.0} & \textbf{25.7} & \textbf{6.2} & \textbf{4.0} & \textbf{54.1} & \textbf{20.7} & \textbf{1.4} & \textbf{3.31} & \textbf{0.35} \\
    \midrule
    LLaVA-v1.5-13B & 31.5 & 20.9 & 53.3 & 14.5 & 8.5 & 50.9 & 37.6 & 4.2 & 3.03 & 0.47 \\
    + $\mathcal{L}_{\text{DPO}}$ with RLAIF-V & 37.1 & \underline{21.2} & \underline{25.8} & \underline{6.3} & \textbf{3.3} & 50.5 & 19.9 & \underline{1.3} & 3.39 & 0.35 \\
    + $\mathcal{L}_{\text{MDPO}}$ with RLAIF-V & 32.8 & 16.9 & 30.3 & 9.1 & \underline{3.4} & \underline{52.6} & \textbf{18.1} & 1.4  & \textbf{3.72} & \textbf{0.30} \\
    + $\mathcal{L}_{\gamma}$ with RLAIF-V-Bias & \textbf{42.1} & \textbf{25.1} & \textbf{23.7} & \textbf{5.9} & 3.5 & \textbf{55.7} & \underline{19.0} & \textbf{1.2} & \underline{3.55} & \underline{0.33} \\
    \bottomrule
    \end{tabular*}
    \caption{\textbf{Main experimental results.} We evaluated our method based on LLaVA-v1.5-7B on bias and hallucination benchmarks, using DPO and MDPO as primary comparisons. Due to differences in training data, model scale, and training strategies, we included additional results for reference. Our method showed an average improvement of 19\% on the bias benchmark and a notable reduction in hallucinations across benchmarks. }
    \label{tab:main_results}
\end{table*}

Table~\ref{tab:main_results} presents our main experimental results, from which we can draw the following conclusions:
(1) \textit{Our method effectively mitigates modality bias in MLLMs.} Compared to the strongest baseline, our approach achieves an average improvement of 19\% on the modality bias benchmark for MLLMs.
(2) \textit{The optimization objective is crucial in addressing modality bias in MLLMs.} Although MDPO is designed for hallucination issues, its approach is similar to ours in addressing language bias.  However, the results on the bias benchmark indicate that MDPO does not effectively mitigate modality bias in MLLMs.
(3) \textit{There is a connection between hallucination and modality bias issues in MLLMs.} In addition to DPO, our enhancements—bias preference optimization data and NaPO—show that our approach reduces hallucination in MLLMs while mitigating modality bias.
These observations underline the effectiveness of our method in addressing both modality bias and hallucination.

\subsection{Variant Analysis}
To better understand the contribution of each module in our method, we selected VLind-Bench, a bias benchmark, and Object HalBench, a hallucination benchmark, as our main evaluation sets. We conducted detailed variant experiments, as shown in Table~\ref{tab:variant_analysis}. The experiments are divided into three groups: full data variant analysis (A), language-bias variant analysis (LB), and vision-bias variant analysis (VB). Here, \textit{w/o} denotes \textit{without}, and \textit{repl.} indicates \textit{replace with}.

\begin{table}[!t]
    \centering\footnotesize
    \scalebox{0.98}{
    \begin{tabular*}{0.48 \textwidth}{@{\extracolsep{\fill}}@{}ll|cccc@{}}
    \toprule
    \multirow{2}{*}{No.} & \multirow{2}{*}{Variant} & \multicolumn{2}{c}{\textbf{VLindBench}} &
    \multicolumn{2}{c}{\textbf{Object HalBench}} \\
    \cmidrule{3-4}
    \cmidrule{5-6}
    ~ & ~ & CB $\uparrow$ & LP $\uparrow$ & CHAIRs $\downarrow$ & CHAIRi $\downarrow$ \\
    \midrule
    A1 & - & \textbf{58.9} & \textbf{44.0} & \textbf{25.7} & \textbf{6.2} \\
    A2 & w/o $\gamma_i$ & 50.0 & 38.2 & 27.7 & 8.0 \\ 
    A3 & repl. DPO & 43.4 & 32.2 & 29.0 & 8.3 \\
    \midrule
    LB1 & w/o VB & 40.4 & 36.4 & 28.0 & 6.4 \\
    LB2 & repl. DPO & 44.7 & 29.9 & 28.3 & 6.8 \\
    \midrule
    VB1 & w/o LB & 62.3 & 31.4 & 26.3 & 7.6 \\
    VB2 & repl. DPO & 55.3 & 28.5 & 27.7 & 7.4 \\
    \bottomrule
    \end{tabular*}
    }
    \caption{\textbf{Variant analysis.} We divided the variant experiments into three groups based on the data scale. Using controlled variable ablations, we conducted a detailed analysis of the contributions of dynamic weighting, NaPO, language-bias data, and vision-bias data to the final results.}
    \label{tab:variant_analysis}
\end{table}

Through observation and analysis of the variant experiment results, we can draw the following conclusions:
(1) \textit{Dynamic weighting effectively balances the optimization magnitude of different losses}. By comparing A1 and A2 in Table~\ref{tab:variant_analysis}, we observe a noticeable decline in overall model performance when dynamic weighting is removed.
(2) \textit{DPO performs poorly when handling noisy data.} By comparing A3, LB2, and VB2 in Table~\ref{tab:variant_analysis} with their default settings, we observe that DPO not only increases hallucination but also degrades the model’s performance on bias-related issues.
(3) \textit{Language-bias data is more effective than vision-bias data in addressing the model’s language prior issues, while vision-bias data better alleviates commonsense bias.} Comparing LB1 and VB1, we observe that LB1 is more effective for language prior issues, whereas VB1 performs better on commonsense bias issues.
(4) \textit{The two types of data exhibit a synergistic effect.} Comparing the full dataset (A2) with LB1 and VB1 individually, we observe that mixed training with both data types enhances the model’s performance on both bias issues while also reducing hallucinations.
(5) \textit{Vision-bias data is more effective in suppressing hallucinations in the model.} This can be understood in two ways: first, vision-bias data reduces the model’s focus on irrelevant details, preventing it from outputting unrelated elements. Secondly, comparing VB1, LB1, and A2, we observe that vision-bias data leads to significantly lower hallucination rates.

\subsection{Further Analysis}

\begin{table}[!t]
    \centering\footnotesize
    \scalebox{0.98}{
    \begin{tabular*}{0.48 \textwidth}{@{\extracolsep{\fill}}@{}l|cccc@{}}
    \toprule
    \multirow{2}{*}{Data} & \multicolumn{2}{c}{\textbf{VLindBench}} &
    \multicolumn{2}{c}{\textbf{Object HalBench}} \\
    \cmidrule{2-3}
    \cmidrule{4-5}
    ~ & CB $\uparrow$ & LP $\uparrow$ & CHAIRs $\downarrow$ & CHAIRi $\downarrow$ \\
    \midrule
    RLAIF-V-Bias & \textbf{58.9} & \textbf{44.0} & \textbf{25.7} & \textbf{6.2} \\
    RLAIF-V & 39.4 & 25.4 & 32.0 & 8.5 \\
    RLAIF-V-Bias (Random) & 51.3 & 40.9 & 28.3 & 7.2 \\
    \bottomrule
    \end{tabular*}
    }
    \caption{\textbf{Quantitative analysis of training data.} We use the same training settings as in the main experiment but vary the amount of training data for comparison. “Random” indicates that, for each batch, we randomly select one type of data from the original, language-bias, or vision-bias data, ensuring the total training data volume matches the original RLAIF-V dataset. The results show that \textit{while reducing data volume leads to performance degradation, our data is more effective than the original data at the same volume.}}
    \label{tab:quantitative_analysis_of_training_data}
\end{table}

\paragraph{Quantitative analysis of training data.}
In this part, we aim to construct a training dataset of equal size to RLAIF-V, while effectively leveraging the advantages of different data types within RLAIF-V-Bias for comparison. 
All experimental settings are consistent with those in the main experiment. 
Specifically, for each batch, we randomly select one type of training data—original, language-bias, or vision-bias—as the current batch’s training data. 
This approach can be viewed as randomly selecting one of the three losses in the final optimization objective (Equation~(\ref{eq:final objective})) for each batch, setting its weight to 1 while setting the others to 0, to enable a quantitative analysis of the training data. 
Observation and analysis of the experimental results in Table~\ref{tab:quantitative_analysis_of_training_data} indicate that \textit{with an equivalent amount of training data, RLAIF-V-Bias (Random) still outperforms RLAIF-V in mitigating hallucination and bias issues, which demonstrate the effective of our motivation.} 
However, the performance of the RLAIF-V-Bias (Random) still falls short compared to the RLAIF-V-Bias.

\begin{table}[!t]
    \centering\footnotesize
    \scalebox{0.98}{
    \begin{tabular*}{0.48 \textwidth}{@{\extracolsep{\fill}}@{}l|cccc@{}}
    \toprule
    \multirow{2}{*}{Margin} & \multicolumn{2}{c}{\textbf{VLindBench}} &
    \multicolumn{2}{c}{\textbf{Object HalBench}} \\
    \cmidrule{2-3}
    \cmidrule{4-5}
    ~ & CB $\uparrow$ & LP $\uparrow$ & CHAIRs $\downarrow$ & CHAIRi $\downarrow$ \\
    \midrule
    LB with $\psi_{\Sigma}$ & 34.8 & 22.2 & 28.7 & 7.1 \\
    LB with $\psi_{\mu}$ (default) & 40.4 & 36.4 & 28.0 & 6.4 \\
    \midrule
    VB with $\psi_{\Sigma}$ (default) & 62.3 & 41.4 & 26.3 & 7.6 \\
    VB with $\psi_{\mu}$ & 52.3 & 29.5 & 28.7 & 7.8 \\
    \bottomrule
    \end{tabular*}
    }
    \caption{\textbf{Relationship between $q$ calculation method and data type.} We used the same hyperparameter settings as in the main experiment and tested various methods for calculating $q$ across different datasets, where the calculation of the $\psi_{\mu}$ and $\psi_{\Sigma}$ is shown in Equation~(\ref{eq:reward_margin}). The results indicate that \textit{using inappropriate noise estimation methods to calculate the noise robustness coefficient leads to performance degradation}.}
    \label{tab:margin_analysis}
\end{table}

\paragraph{Noise evaluation metric analysis.}
In this chapter, we analyze the impact of different noise evaluation methods on the final results by applying distinct noise assessment methods to various data types. 
All experimental settings and hyperparameters remain consistent with those in the main experiment. 
Specifically, for the scale factor $\alpha$ in Equation~(\ref{eq:dynq}), we use the default settings: $\alpha = 0.5$ for reward margins $\psi_{\mu}$ calculated using average logp, and $\alpha = 0.01$ for reward margins $\psi_{\Sigma}$ calculated using logp, to ensure comparability of results.
Observing the experimental results in Table~\ref{tab:margin_analysis}, we see that \textit{using inappropriate noise evaluation methods leads to a sharp decline in model performance}. 
This may be because incorrect noise assessment increases the gradient for noisy data in the loss function while decreasing it for non-noisy data. 
As a result, the model may overfit to noisy data and overlook high-quality data during optimization. 
This outcome also supports the pattern we summarized in Section~\ref{sec:napo} based on our analysis of Figure~\ref{fig:dist}.




\section{Related Work}
\label{sec:related}

\subsection{Modality Bias in MLLMs}
Modality bias occurs when a model overly relies on one modality or prior knowledge, neglecting other relevant modalities~\cite{guo2023modality, chen2024quantifying}. 
This leads to a focus on incorrect information and weakens generalization, manifesting as language bias~\cite{agrawal2018don, abbasnejad2020counterfactual, zhu2020overcoming, cadene2019rubi} or vision bias~\cite{gupta2022swapmix, si2022language}. 
Related to the hallucination problem~\cite{zhang2024debiasing}, modality bias has been extensively studied in VQA tasks~\cite{guo2023modality, chen2024quantifying, agrawal2018don, niu2021counterfactual, zhu2020overcoming, cadene2019rubi, gokhale2020mutant}. 
However, most prior work has focused on balanced datasets and complex training strategies, which do not generalize well to MLLMs.
Recent studies addressing modality bias in MLLMs include benchmarks for evaluating bias~\cite{lee2024vlind}, datasets for assessing modality bias~\cite{chen2024quantifying}, and methods like contrastive decoding to reduce language priors~\cite{zhang2024debiasing}. 
While MDPO~\cite{wang2024mdpo} takes an approach similar to ours, it targets hallucination by lowering the probability of preferred samples when images are absent, differing from our method that directly reduces the probability of biased responses without images.
Additionally, works such as \cite{wang2024can, chuang2023debiasing, seth2023dear, zhang2024vlbiasbench} address social and group biases in vision-language models like CLIP. 
Although similar in theme, these studies primarily focus on social biases, which diverges from our focus on modality bias in MLLMs.

\subsection{Preference Optimization}
Preference optimization algorithms enhance LLMs by aligning them with human values~\cite{ouyang2022training}. While RLHF\cite{christiano2017deep, bai2022training, touvron2023llama, ouyang2022training, chowdhury2024provably, chen2024noise} is effective, its complexity has led to exploration of simpler alternatives. RAFT~\cite{dong2023raft} selects optimal samples with existing reward models, RRHF~\cite{yuan2023rrhf} uses a simpler ranking loss, and DPO~\cite{rafailov2024direct} employs a preference-based loss for improved stability. SLiC-HF~\cite{zhao2023slic}, KTO~\cite{ethayarajh2024kto}, RSO~\cite{liu2023statistical}, and ORPO~\cite{hong2024orpo} focus on preference calibration and efficient modeling. $\beta$-DPO~\cite{wu2024beta} dynamically adjusts $\beta$ based on data distribution. 
With the increasing scale of models and training datasets, obtaining human-annotated gold data has become increasingly challenging~\cite{cao2024towards}. Recently, LLM-synthesized data~\cite{tan2024large} has garnered significant attention. However, synthesized data inevitably introduces noise. Current research primarily focuses on leveraging automatically constructed data~\cite{tan2024large} and AI-based feedback~\cite{gu2024survey, li2024generation} to address existing challenges, while relatively little attention has been paid to enhancing the noise robustness of training methods.
In contrast, NaPO integrates noise-robust MAE into DPO and dynamically adjusts noise robustness based on data noise, enhancing stability and resilience in noisy environments.

\section{Conclusion}
\label{sec:conclusion}

We introduce RLAIF-V-Bias, a dataset designed to optimize preferences in MLLMs and reduce modality bias by including both language and visual bias data. To handle noise in automatically constructed data, we propose NaPO, which combines BCE loss from DPO with noise-robust MAE loss using a negative Box-Cox transformation, allowing dynamic noise detection and robustness adjustment. 
Experimental results demonstrate that our method effectively mitigates bias and significantly reduces hallucinations in the model outputs. Compared to DPO, NaPO exhibits stronger resilience to noise in automatically constructed data. 
However, two clouds still linger behind our research: \textit{whether NaPO can robustly handle noise originating from LLM-synthesized data in broader contexts}, and \textit{whether bias is always harmful}.

\section*{Acknowledgement}
\label{sec:acknowledgement}

We would like to thank the anonymous reviewers for their comments. This work is supported by the National Natural Science Foundation of China (No.62406319), the Youth Innovation Promotion Association of CAS (No.2021153), and the Postdoctoral Fellowship Program of CPSF (No.GZC20232968).

{\small
\bibliographystyle{ieeenat_fullname}
\bibliography{11_references}
}

\clearpage \appendix 


\section{Additional Experiments}
\label{sec:additional_experiments}

\subsection{Hyperparameter Analysis of the NaPO}

\begin{figure*}
\centering
\includegraphics[width=1\textwidth]{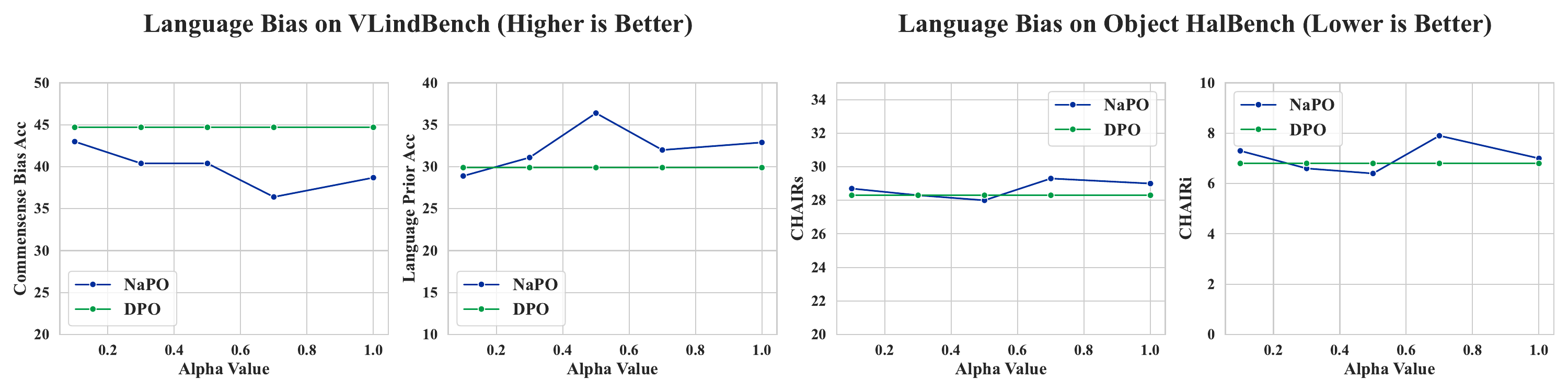}
\caption{\textbf{Hyperparameter analysis on language-biasd data.} The chart illustrates the results of the model on VLindBench and Object HalBench when training on language-biased data with different $\alpha$ values in NaPO. We observed that the model achieves better performance across all four metrics when $\alpha$ is set to 0.5.} 
\label{fig:lbhyper}
\end{figure*}

\begin{figure*}
\centering
\includegraphics[width=1\textwidth]{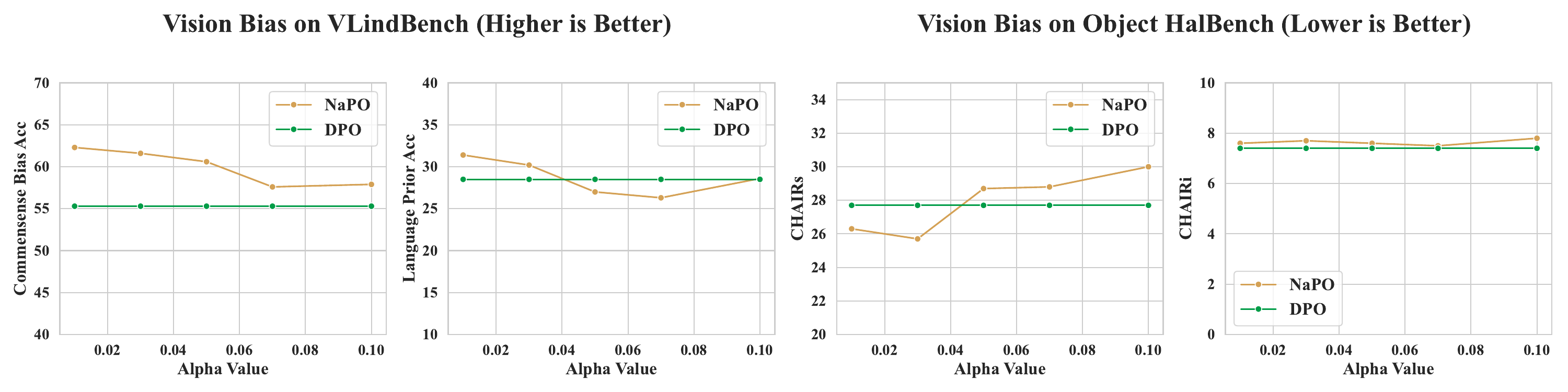}
\caption{\textbf{Hyperparameter analysis on vision-biasd data.} The chart illustrates the results of the model on VLindBench and Object HalBench when training on vision-biased data with different $\alpha$ values in NaPO. We observed that the model’s performance gradually decreases as the $\alpha$ value increases.} 
\label{fig:vbhyper}
\end{figure*}

\begin{figure}
\centering
\includegraphics[width=0.48 \textwidth,height=0.25\textwidth]{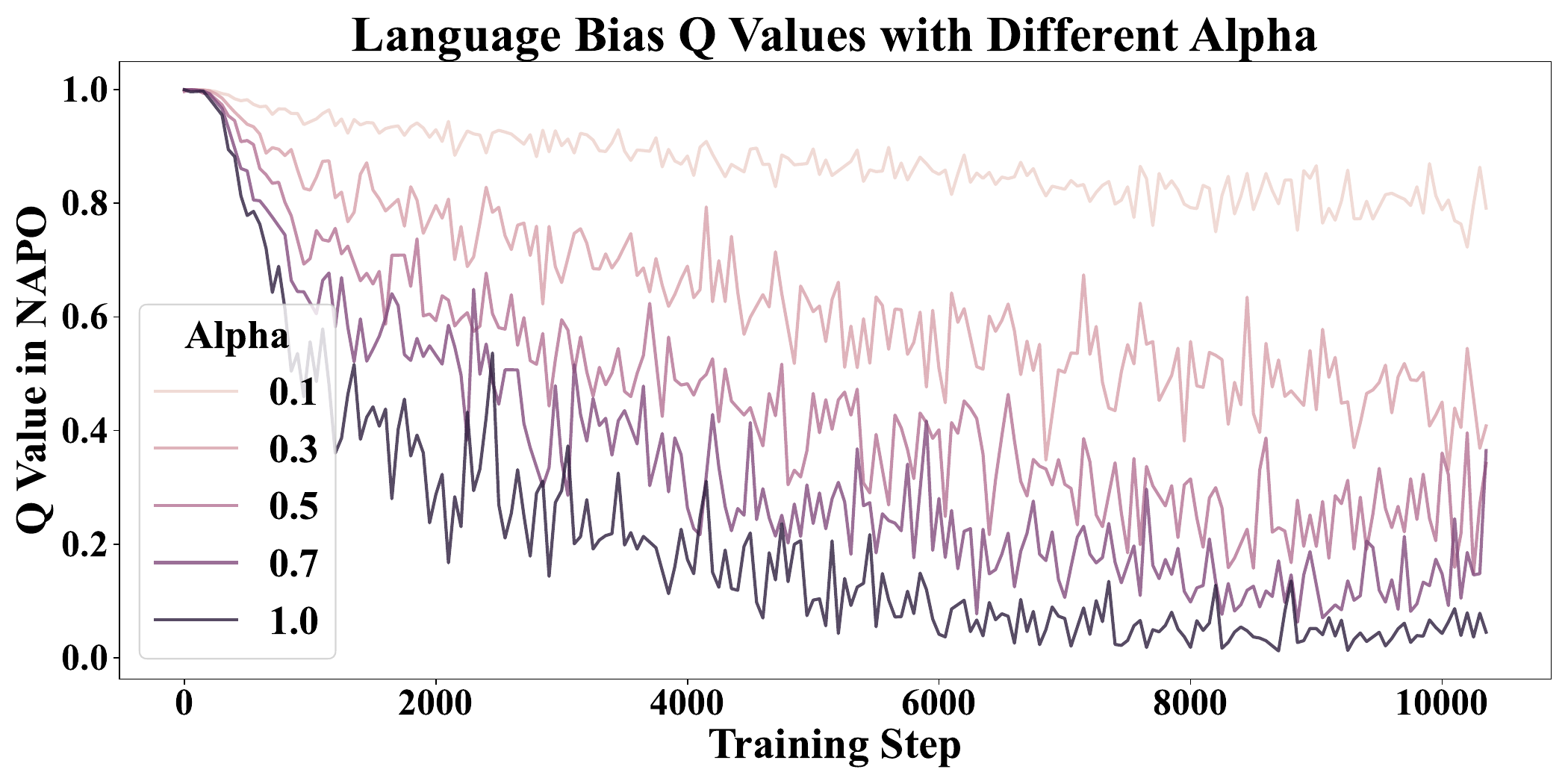}
\caption{\textbf{The trend of $q$ during training.} The figure illustrates the variation of $q$ during training under different $\alpha$ values.  As $\alpha$ increases, $q$ exhibits larger fluctuations and decreases at a faster rate.  In contrast, smaller $\alpha$ values result in more stable changes in $q$, with a slower and more consistent decline.} 
\label{fig:qvalues}
\end{figure}

In this section, we detail the strategy for selecting the scaling parameter $\alpha$ and evaluate the model’s performance under various $\alpha$ values across different datasets.

Firstly, as shown in Figure~\ref{fig:dist}, the numerical range of the logp margin is typically larger compared to the avg logp margin. To ensure effective value scaling, we leverage the sigmoid function, which is most sensitive to changes within the range of $-2 \sim 2$. To achieve this, we scale the margin values using the parameter $\alpha$. Specifically, for logp margins, $\alpha$ is selected within the range $[0.01, 0.1]$, while for avg logp margins, $\alpha$ is chosen from the range $[0.1, 1.0]$.

Secondly, Figures~\ref{fig:lbhyper} and~\ref{fig:vbhyper} illustrate the model’s performance on VLindBench and Object HalBench when trained on language-biased and vision-biased data, respectively. The results demonstrate that selecting an appropriate $\alpha$ allows NaPO to estimate the noise-robust coefficient $q$ more effectively, leading to improved model performance. In contrast, using an unsuitable $\alpha$ value can result in suboptimal $q$ estimation and degrade performance.

Finally, in Figure~\ref{fig:qvalues}, we analyze the variation of $q$ during training under different $\alpha$ settings. We observe that higher $\alpha$ values amplify the fluctuations of $q$ during training and cause $q$ to decrease more rapidly. Conversely, smaller $\alpha$ values stabilize $q$, keeping it at relatively higher values throughout training. This analysis underscores the importance of carefully selecting $\alpha$ to balance robustness and adaptability during training.

\subsection{NaPO with RLAIF-V}

\begin{table}[!t]
    \centering\footnotesize
    \scalebox{0.98}{
    \begin{tabular*}{0.48 \textwidth}{@{\extracolsep{\fill}}@{}l|cccc@{}}
    \toprule
    \multirow{2}{*}{Loss and Data} & \multicolumn{2}{c}{\textbf{VLindBench}} &
    \multicolumn{2}{c}{\textbf{Object HalBench}} \\
    \cmidrule{2-3}
    \cmidrule{4-5}
    ~ & CB $\uparrow$ & LP $\uparrow$ & CHAIRs $\downarrow$ & CHAIRi $\downarrow$ \\
    \midrule
    $\mathcal{L}_{\text{DPO}}$ with RLAIF-V & 39.4 & \textbf{25.4} & 32.0 & 8.5 \\
    $\mathcal{L}_{\text{MDPO}}$ with RLAIF-V & 0.3 & 0.4 & 35.3 & 10.5 \\
    $\mathcal{L}_{\text{NaPO}}$ with RLAIF-V & \textbf{48.3} & 22.3 & \textbf{26.7} & \textbf{7.5} \\
    \bottomrule
    \end{tabular*}
    }
    \caption{\textbf{NaPO with RLAIF-V.} We tested the effectiveness of NaPO on the RLAIF-V and found that the results of NaPO outperform those of DPO on the RLAIF-V.}
    \label{tab:napo_with_rlaifv}
\end{table}

\begin{table}[!t]
    \centering\footnotesize
    \scalebox{0.98}{
    \begin{tabular*}{0.48 \textwidth}{@{\extracolsep{\fill}}@{}l|cccc@{}}
    \toprule
    \multirow{2}{*}{Loss} & \multicolumn{2}{c}{\textbf{VLindBench}} &
    \multicolumn{2}{c}{\textbf{Object HalBench}} \\
    \cmidrule{2-3}
    \cmidrule{4-5}
    ~ & CB $\uparrow$ & LP $\uparrow$ & CHAIRs $\downarrow$ & CHAIRi $\downarrow$ \\
    \midrule
    $\mathcal{L}_{\gamma}$ & \textbf{58.9} & 44.0 & \textbf{25.7} & \textbf{6.2} \\
    w/o $\gamma_i$ + repl. $\mathcal{L}_{\text{NaPO}}$ & 54.0 & \textbf{47.8} & 27.3 & 7.0 \\
    repl. $\mathcal{L}_{\text{NaPO}}$ & 21.9 & 21.1 & 36.7 & 9.2 \\
    \bottomrule
    \end{tabular*}
    }
    \caption{\textbf{Replace with NaPO.} We found that replacing DPO in $\mathcal{L}_{\gamma}$ with NaPO leads to a certain degree of performance degradation. Moreover, utilizing $\gamma_i$ to balance the loss weights in this case causes the model performance to decline sharply. Therefore, dynamic weight balancing may not be suitable for all scenarios.}
    \label{tab:napo_with_all}
\end{table}


To evaluate the effectiveness of NaPO on the original dataset, we conducted experiments using the same default settings as the main experiments. Specifically, we used $\log p$ to estimate the noise-robust coefficient $q$, and, consistent with the main experiments, we set $\alpha = 0.01$. The experimental results are shown in Table~\ref{tab:napo_with_rlaifv}. 
From Table~\ref{tab:napo_with_rlaifv}, we observed that NaPO outperforms DPO and MDPO across most metrics. This observation prompted the question: \textit{would replacing DPO with NaPO in $\mathcal{L}_{\gamma}$ lead to further performance improvements}? 
To explore this, we replaced DPO with NaPO in $\mathcal{L}_{\gamma}$, and the results are presented in Table~\ref{tab:napo_with_all}. Surprisingly, this replacement resulted in a performance drop. Moreover, under these conditions, employing dynamic weight balancing with $\gamma_i$ caused the model to collapse. This indicates that dynamic weight balancing with $\gamma_i$ may not be suitable for all scenarios. We leave the detail discussion of these issues for future work.

\section{Data Construction and Analysis}
\label{sec:data_construction_and_analysis}

\subsection{Data Construction}
We use the LLaVA-v1.5-7B model to construct our dataset, employing inference hyperparameters of temperature \texttt{t=0}, \texttt{num\_beams=3}, and \texttt{max\_new\_tokens=1024}. No additional prompts are utilized to guide the model. For language-biased responses, we retain only the textual information from the questions to prompt the model’s answers, whereas for vision-biased responses, we exclusively preserve visual information from the questions to generate the model’s replies.
Additionally, we do not employ any explicit data-filtering strategies. Our NaPO approach can instead be viewed as a soft data-selection method, dynamically adjusting the optimization strength by adaptively controlling the noise robustness coefficient \texttt{q}.

\subsection{Data Analysis}
\label{sec:data_observation_and_analysis}
From the observation of Figure~\ref{fig:dist}, we note that in language-biased responses, noise-free (biased) data exhibit a higher avg LogP margin compared to noisy (unbiased) data. Similarly, in vision-biased responses, noise-free (biased) data show a higher LogP margin than noisy (unbiased) data.

To better understand this phenomenon, we must clarify the main distinction between LogP and avg LogP. LogP is highly sensitive to response length, meaning that variations in length can significantly influence its value. In contrast, avg LogP normalizes by response length, reducing the impact of length differences and making it more reflective of semantic consistency.

To explore this further, we computed the average length margin of different types of sampled data, as shown in Table~\ref{tab:length margin}. The results indicate that the length margin between noisy and noise-free data in language-biased responses is minimal, suggesting that avg LogP is more effective in distinguishing noise from noise-free samples in this case. Conversely, the substantial length margin between noisy and noise-free data in vision-biased responses amplifies the influence of length sensitivity, making LogP a better metric for identifying noise in vision-biased responses.

\begin{table}[ht]
    \centering\footnotesize
    \setlength{\tabcolsep}{5pt}
    \scalebox{0.98}{
    \begin{tabular*}{0.38\textwidth}
    {@{\extracolsep{\fill}}@{}c|cc@{}}
    \toprule
    Mean of Length Margin & Language Bias & Vision Bias \\
    \midrule
    Biased (noise-free) & 60.83 & -199.20 \\
    Unbiased (noisy) & 65.57 & 199.61 \\
    \bottomrule
    \end{tabular*}}
    \caption{\textbf{Analysis of noise and length margin.} We observe that vision-biased responses have significant length margin differences between noise and non-noise data, while language-biased responses show minimal length margin differences between them.}
    \label{tab:length margin}
\end{table}

The differences in length margins between noisy and noise-free data can be attributed to the vision-biased generation process.  
In this process, textual instructions are masked, prompting the model to rely primarily on image content.  
This often results in image captions, which are classified as noisy data when responding to descriptive prompts like “Describe the objects in the image in detail,” as they provide general rather than detailed descriptions.  
Conversely, for question-answering prompts such as “What is the person in the image wearing? ” the responses are typically concise and specific, relying directly on visual information.  
These are categorized as noise-free data.  
Therefore, the length margin for noise-free data is typically very small, whereas that for noisy data is significantly larger.

\section{Case Study}
\begin{figure*}
\centering
\includegraphics[width=1\textwidth]{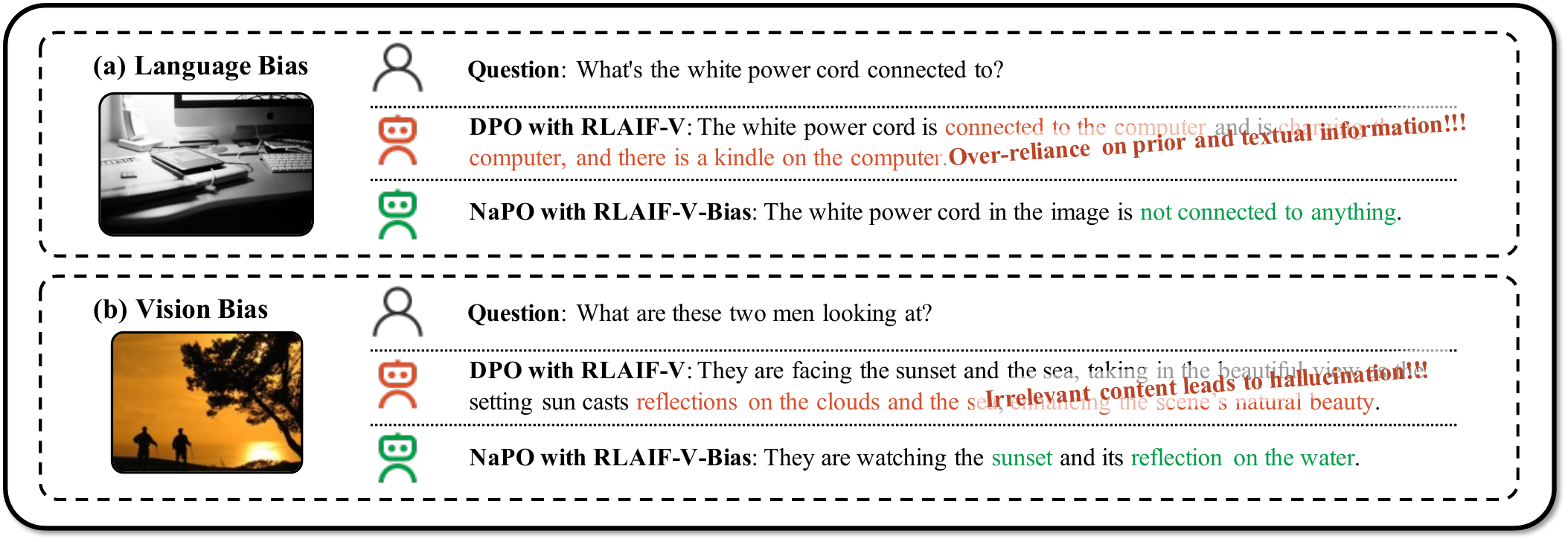}
\caption{\textbf{Case study.} Models trained on general datasets over-rely on prior knowledge and specific modalities. In example (a), the model assumes the power cord is connected to a computer, though it’s visibly disconnected. In example (b), irrelevant details lead to hallucinations of clouds and the sea, despite neither being visible.} 
\label{fig:case}
\end{figure*}

By observing the comparison in Figure~\ref{fig:case}, it becomes evident that models trained on general datasets often rely too heavily on prior knowledge and specific modalities. In example (a), the model assumes that the power cord should be connected to a computer, whereas a closer inspection shows that it isn’t connected to anything in the image. In example (b), the model includes excessive, irrelevant details, resulting in hallucinations of objects like clouds and the sea, even though no clear clouds or ocean are visible in the image.

\end{document}